\documentclass[11pt]{article}

\usepackage[preprint]{acl}
\usepackage{dblfloatfix} 
\usepackage{times}
\usepackage{latexsym}
\usepackage{amsmath}
\usepackage[T1]{fontenc}

\usepackage[utf8]{inputenc}

\usepackage{microtype}
\usepackage{color}
\usepackage{inconsolata}

\usepackage{graphicx}
\usepackage{booktabs}
\usepackage{multirow}
\usepackage{subcaption} 
\usepackage[most]{tcolorbox}
\tcbuselibrary{listings,breakable, skins}

\usepackage[ruled,vlined]{algorithm2e}
\usepackage{xcolor}

\title{Experience Funnel: A State–Policy Alternating Loop for Self-Evolving Agents}
\author{
\textbf{Wenbo Gao\textsuperscript{1,2}},
\textbf{Zhaomou Song\textsuperscript{2}},
\textbf{Zhiyuan Ji\textsuperscript{2,3}},
\textbf{Renxi Liu\textsuperscript{2}},
\textbf{Xing Li\textsuperscript{2}},
\textbf{Xianzhi Yu\textsuperscript{2}},\\
\textbf{Xiaoguang Li\textsuperscript{2}},
\textbf{James Chung-wai CHEUNG\textsuperscript{1}},
\textbf{Weizhe Lin\textsuperscript{2}
    \thanks{Corresponding author.}},
\textbf{Yaoyuan Wang\textsuperscript{2}}
\\
\textsuperscript{1}The Hong Kong Polytechnic University,
\textsuperscript{2}Huawei
\textsuperscript{3}Renmin University of China\\
\texttt{wenbo.gao@connect.polyu.hk}~~~~\texttt{linweizhe1@huawei.com}
}
\usepackage{amsmath} 
\usepackage{amssymb}
\usepackage{cleveref}
\usepackage{svg}
\begin{document}
\maketitle

\begin{abstract}
Autonomous agents powered by large language models (LLMs) continuously accumulate experience through interaction, creating an opportunity to improve future behavior through self-evolution. A fundamental challenge is how to transform abundant, task-specific interaction experience into reusable model competence without sacrificing the ability to adapt rapidly to newly observed evidence. Explicit textual states, such as skills and agent harnesses, provide fast, human-readable and editable adaptation, but incur persistent dependence on external context; parametric policies provide compact and reusable competence, but are substantially slower to update. We present \textit{Experience Funnel}, a self-evolving framework that couples fast state adaptation with slow policy consolidation in an alternating loop. Interaction trajectories are first distilled into an explicit textual state, where newly acquired experience can be rapidly incorporated and validated. The framework then selectively identifies state-enabled behavior that remains useful across state revisions and consolidates it into the policy through transition-aware distillation. The updated state--policy pair subsequently generates new rollouts, providing fresh evidence for the next round of state adaptation and policy consolidation. Experiments across diverse agent benchmarks show that \textit{Experience Funnel} consistently improves agent capability over state-only evolution and policy-internalization approaches, while progressively converting useful explicit experience into autonomous policy competence.
\end{abstract}

\section{Introduction}

Language agents deployed across repeated interaction episodes continually generate experience that can be reused to improve future behavior rather than solving every task from scratch. Successful trajectories reveal effective strategies and procedures, whereas failures expose incorrect decisions, missing knowledge, and capability gaps. Existing self-evolving agents preserve such experience through two complementary forms of adaptation. One is to externalize experience as textual states---including memories, skills, procedural instructions, and agent harnesses---that can be rapidly revised without modifying model parameters
\cite{
shinn2023reflexion,
zhao2024expel,
wang2024voyager,
cai2025ell,
zhang2025ace,
yang2026skillopt}.
The other is to internalize useful experience into the parametric policy through reinforcement learning or distillation, converting interaction-derived guidance into autonomous model behavior
\cite{
yu2026selfconsolidation,
yang2026opid,
wang2026skillsd,
lu2026skill0}.
These two representations naturally operate at different adaptation timescales: explicit textual state provides a fast, human-readable and editable interface for newly acquired experience, whereas policy parameters provide a slower but more compact and persistent substrate for reusable competence.

Neither representation alone, however, provides a satisfactory endpoint for continual self-evolution. Indefinitely accumulating textual experience increases retrieval and context cost, introduces redundant or conflicting guidance, and requires increasingly complex memory and skill management
\cite{
liu2024lostmiddle,
yu2026selfconsolidation,
lu2026skill0,
zhang2026memskill,
lin2026museautoskill,
xu2026hyperskill}.
Conversely, immediately absorbing all observed experience into model parameters risks consolidating noisy, sample-specific, or already mastered behavior. Moreover, the value of explicit experience is policy dependent: guidance that is useful to the current policy may become redundant, complementary, or even conflicting after the policy itself improves
\cite{
xia2026skillrl,
he2026reskill,
chen2026harnessforge,
lu2026skill0}.
The central challenge is therefore not merely how to accumulate more experience, but how to progressively transform interaction experience into reusable model competence while preserving a fast and editable mechanism for continued adaptation.

Recent work has addressed individual parts of this problem. Skill- and memory-evolution methods maintain persistent textual guidance and revise it from interaction feedback
\cite{
zhang2025ace,
yang2026skillopt,
he2026reskill}.
Skill-conditioned reinforcement learning and self-distillation instead transfer textual guidance into policy behavior
\cite{
xia2026skillrl,
wang2026skillsd,
lu2026skill0,
lin2026skillc}.
OPID extracts hindsight skills from on-policy trajectories and attributes their behavioral effects at the token level, while SEED further couples trajectory collection with evolving skill analysis
\cite{
yang2026opid,
wu2026seed}.
Other approaches explicitly recognize the dependence between textual guidance and policy behavior: SkillRL recursively evolves a skill library together with reinforcement learning, ReSkill evaluates skill revisions under the evolving policy, and HarnessForge jointly adapts agent harnesses and policy parameters
\cite{
xia2026skillrl,
he2026reskill,
chen2026harnessforge}.
Together, these studies establish the importance of both explicit experience and parametric adaptation. Yet they leave open a broader experience-management question: how should interaction experience be progressively filtered from raw trajectories into editable state, selectively consolidated into long-term policy competence, and subsequently reconsidered as the evolving policy generates new behavior?

We address this question with \emph{Experience Funnel}, a self-evolving framework that couples fast state adaptation with slow policy consolidation in an alternating loop. At each evolution round, trajectories generated by the current state--policy pair are first aggregated into a task-specific textual state that summarizes recurring procedures, failure modes, and corrective strategies. Candidate state updates are evaluated on held-out interactions so that newly observed experience can be rapidly incorporated while sample-specific or unreliable edits are filtered out. The resulting explicit state therefore serves as an intermediate experience representation: sufficiently flexible for rapid adaptation, yet structured enough to expose reusable behavioral knowledge for subsequent consolidation.

Not all state-enabled behavior, however, should become parametric competence. We therefore introduce \emph{Transition-Aware Skill Distillation}, which compares state-free execution with behavior conditioned on the previous and updated states to identify experience that is newly useful, persistently useful, regressive, or inactive. Only behavior that remains beneficial under the evolving state is selected for consolidation, while token-level state-conditioned contrasts localize the decisions attributable to explicit experience. A state-conditioned teacher then transfers these selected behavioral effects into a state-free policy. Once updated, the policy is redeployed together with the current state and generates a new distribution of interaction trajectories. These rollouts provide fresh evidence for the next round of state adaptation, allowing complementary experience to be retained or refined and experience already absorbed by the policy to become unnecessary. 

We evaluate \emph{Experience Funnel} across question answering, embodied interaction, and web navigation. Experiments show that alternating state adaptation and policy consolidation consistently improves agent performance over state-only evolution and policy-internalization approaches. State-free policy performance increases across accepted consolidation rounds, demonstrating that useful explicit experience is progressively converted into autonomous competence, while the remaining textual state preserves complementary guidance that has not yet been internalized. These results support an experience-centric view of self-evolution in which rapid explicit adaptation and slower parametric consolidation play distinct but complementary roles.

\section{Experience Funnel for Self-Evolving Agents}
\label{sec:method}

We introduce \emph{Experience Funnel}, a self-evolving framework that enables agents to continuously learn from interaction experience through complementary state and policy representations. Our approach is realized from two perspectives: state--policy alternating evolution (Sec.~\ref{sec:loop}) and transition-aware on-policy distillation (Sec.~\ref{sec:transition_distillation}). For continual experience accumulation, explicit textual states enable fast and editable adaptation, but their continual growth leads to increasing context and retrieval overhead. To address this, we introduce a \emph{State--Policy Alternating Loop}, where newly acquired experience is first organized in the task-specific state, while reusable experience is progressively consolidated into the long-term policy, allowing the two experience representations to evolve jointly. For policy consolidation, directly training on all state-conditioned trajectories may introduce redundant or weakly relevant supervision. To address this, we propose \emph{Transition-Aware On-Policy Distillation}, which selects informative trajectories according to cross-state behavioral transitions and further identifies state-responsive tokens for focused policy optimization.

\begin{figure*}[t]
    \centering
    \includegraphics[width=\textwidth]{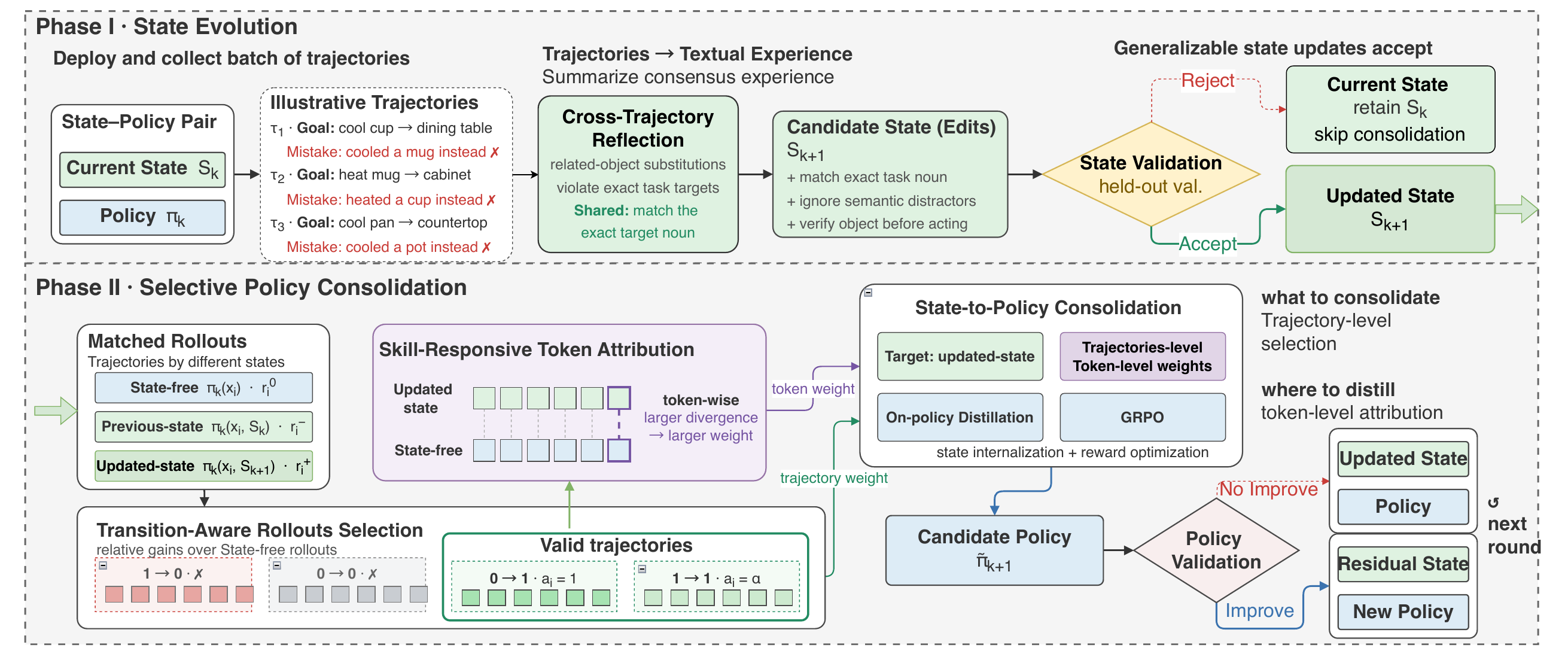}
    \caption{
    Overview of the \emph{State--Policy Consolidation Loop}.
    Phase I aggregates recurring patterns across interaction trajectories into
    a validated textual state update. Phase II identifies state-enabled behavior
    through transition-aware rollout selection and token-level attribution, and
    selectively consolidates it into the state-free policy. The updated
    state--policy pair then returns to deployment and initiates the next
    evolution round.
    }
    \label{framework}
\end{figure*}

\subsection{State--Policy Alternating Loop}
\label{sec:loop}

\paragraph{State Evolution.}
During deployment, the current state--policy pair accumulates interaction
trajectories. Once sufficient trajectories have been collected, we freeze them
as a reflection batch and jointly identify recurring behavioral patterns across
interactions rather than deriving corrections from individual examples. These
patterns are summarized into high-level textual experience, including reusable
procedures, common failure modes, and corrective strategies, and represented as
candidate edits to the current state.

Inspired by validation-gated textual skill optimization
\cite{yang2026skillopt,shen2026skilloptlite}, candidate updates are evaluated
on a separate held-out validation set under the current policy and accepted only
when they improve the environment-level outcome. Cross-trajectory reflection
therefore extracts reusable experience, while held-out validation filters
sample-specific or unreliable updates.

\paragraph{Policy Consolidation and State Revision.}
An accepted state update improves the current agent but also increases its
dependence on deployment-time context. We therefore selectively consolidate its
useful behavioral effect into the parametric policy
\cite{yu2026selfconsolidation,lu2026skill0,ye2026opcd,wang2026skillsd}, with
the selection and training objective described in
Sec.~\ref{sec:transition_distillation}.

After consolidation, the updated policy $\pi_{k+1}$ returns to deployment with
the current textual state $S_{k+1}$. Subsequent interaction trajectories provide
both new experience and delayed evidence about the utility of existing guidance:
complementary state components are retained or refined, redundant components can
be retired, and unresolved failures motivate new state updates
\cite{he2026reskill,lu2026skill0,tu2026ucob}. The resulting evolution is
\begin{equation}
(S_k,\pi_k)
\rightarrow
S_{k+1}
\rightarrow
\pi_{k+1}
\rightarrow
S_{k+2}
\rightarrow
\pi_{k+2}
\rightarrow \cdots .
\end{equation}
This policy-conditioned revision prevents indefinite accumulation of textual
guidance while preserving experience that remains useful to the evolving policy.

\subsection{Transition-Aware Skill Distillation}
\label{sec:transition_distillation}

A validated state does not make every state-conditioned behavior suitable for
consolidation, since the state-free policy may already perform part of that
behavior correctly. We therefore use trajectory-level transitions to identify
behavior that the updated state adds to the current policy, and token-level
contrasts to localize the decisions attributable to state conditioning.

\paragraph{Transition-Aware Rollout Selection.}
For each interaction instance $x_i$, we evaluate the current policy $\pi_k$
without textual state, with the previous state $S_k$, and with the validated
updated state $S_{k+1}$
\cite{lin2026skillc,he2026reskill,tu2026ucob}:
\begin{equation}
\begin{aligned}
r_i^{0} &= R(\pi_k,x_i),\\
r_i^{-} &= R(\pi_k,x_i,S_k),\\
r_i^{+} &= R(\pi_k,x_i,S_{k+1}),
\end{aligned}
\end{equation}
where $R(\cdot)$ denotes the environment outcome. Their behavioral utility
relative to state-free execution is
\begin{equation}
b_i^{-}=\mathbb{I}[r_i^{-}>r_i^{0}],
\qquad
b_i^{+}=\mathbb{I}[r_i^{+}>r_i^{0}].
\end{equation}

The transition $(b_i^{-},b_i^{+})$ distinguishes newly useful $(0,1)$ and
persistently useful $(1,1)$ behavior from regressive $(1,0)$ and inactive
$(0,0)$ behavior. We retain the first two categories for consolidation using
\begin{equation}
a_i=
\begin{cases}
1, & (b_i^{-},b_i^{+})=(0,1),\\
\alpha, & (b_i^{-},b_i^{+})=(1,1),\\
0, & \text{otherwise},
\end{cases}
\end{equation}
where $\alpha$ controls the relative contribution of persistently useful
behavior.

\paragraph{State-Responsive Token Attribution.}
Trajectory-level selection identifies rollouts that benefit from the updated
state, but not which decisions within them are attributable to state
conditioning. For each selected trajectory and token position $t$, we therefore
compare the frozen policy $\pi_k$ under identical prefixes with and without the
accepted state
\cite{yang2026opid,wang2026skillsd,tu2026ucob}:
\begin{equation}
\begin{aligned}
p_{i,t}^{+}
&=
\pi_k\!\left(
\cdot\mid x_i,S_{k+1},y_{i,<t}
\right),\\
p_{i,t}^{0}
&=
\pi_k\!\left(
\cdot\mid x_i,y_{i,<t}
\right).
\end{aligned}
\end{equation}
Because the two branches share the same model parameters and differ only in
textual-state conditioning, their predictive difference isolates the local
behavioral response to the accepted state.

We quantify this response using Jensen--Shannon divergence,
\begin{equation}
d_{i,t}
=
D_{\mathrm{JS}}
\!\left(
p_{i,t}^{+}
\,\Vert\,
p_{i,t}^{0}
\right),
\end{equation}
and normalize it within each trajectory:
\begin{equation}
w_{i,t}
=
\frac{d_{i,t}}
{\frac{1}{T_i}\sum_{t'=1}^{T_i}d_{i,t'}+\epsilon}.
\end{equation}
Thus, $a_i$ determines which rollouts contribute to consolidation, while
$w_{i,t}$ emphasizes decisions that are most responsive to the accepted state.

\paragraph{Policy Consolidation.}
Let $\pi_\theta$ be a trainable policy initialized from $\pi_k$. The
state-conditioned branch $p_{i,t}^{+}$ provides privileged supervision, while
$\pi_\theta$ receives only the state-free context
\cite{ye2026opcd,wang2026skillsd}. We optimize
\begin{equation}
\mathcal{L}_{\mathrm{distill}}
=
\frac{
\displaystyle
\sum_{i,t}
a_iw_{i,t}
D_{\mathrm{KL}}
\!\left(
\pi_\theta(\cdot\mid x_i,y_{i,<t})
\,\Vert\,
p_{i,t}^{+}
\right)
}{
\displaystyle
\sum_{i,t}a_iw_{i,t}
},
\end{equation}
which concentrates distillation on behavior that is both outcome-relevant and
responsive to state conditioning.

We combine selective distillation with state-free reward optimization
\cite{wang2026skillsd,lu2026sdar}:
\begin{equation}
\mathcal{L}_{\mathrm{consolidate}}
=
\mathcal{L}_{\mathrm{RL}}^{\mathrm{state\text{-}free}}
+
\lambda_{\mathrm{distill}}
\mathcal{L}_{\mathrm{distill}}.
\end{equation}
The reward objective improves autonomous behavior from environment feedback,
whereas distillation internalizes behavior enabled by the validated state. The
candidate policy $\tilde{\pi}_{k+1}$ is committed only when its state-free
performance satisfies the validation criterion, after which it returns to
deployment and initiates the next state--policy evolution round.

\section{Experiments}

\subsection{Experimental Setup}

\paragraph{Task and evaluation.}
We first study state--policy consolidation on SearchQA using Qwen3.5-4B as the evolving deployment policy and a fixed Qwen3.5-27B model as the OPD teacher. Hard answer accuracy (ACC) is the primary evaluation metric, while the environment reward is used for state validation and experience selection.

\paragraph{Implementation and hardware.}
All experiments are conducted on 910B3 Ascend NPUs. Unless otherwise stated, the evolving policy and OPD teacher are deployed within the same experimental environment, with the Qwen3.5-4B model serving as the evolving deployment policy and Qwen3.5-27B as the fixed teacher model. The same hardware and software configuration is used throughout the state--policy evolution process to ensure consistent comparisons across evolution rounds.

\paragraph{Data protocol.}
We strictly separate data according to their roles in the evolution loop. Training examples are used for experience collection, state revision, cross-version transition labeling, and OPD. The validation split is used only to accept state revisions and select policy checkpoints. The test split is reserved exclusively for final reporting and does not affect any subsequent state or policy update.

\paragraph{Evolution protocol.}
We run five state-evolution rounds. The candidate state revisions in rounds 1 and 4 pass the validation gate and trigger policy consolidation, whereas rounds 2, 3, and 5 are rejected and retain the preceding state--policy pair. We therefore report policy updates only at accepted rounds. Unless otherwise stated, each trained configuration corresponds to a single run; repeated-run uncertainty is reported for the final comparison.

\paragraph{Cross-environment evaluation.}
We additionally report available results on ALFWorld and WebShop to examine whether the same state--policy pattern transfers across interactive domains. Because the current experiments use different OPD budgets across environments, cross-environment averages are treated as descriptive summaries rather than controlled comparisons under an identical training budget.

\subsection{Overall Performance}
\label{sec:overall}

Table~\ref{tab:main-results} compares our State--Policy Consolidation Loop
with representative approaches based on explicit-state evolution, policy
internalization, and state--policy co-evolution. Our method achieves the best
performance across all three environments, reaching 62.4\% on SearchQA,
67.9\% on ALFWorld, and 42.4\% on WebShop, with the highest average score of
57.6\%.

\begin{table*}[t]
\centering
\scriptsize
\setlength{\tabcolsep}{8pt}
\caption{
Overall performance across agent environments.
All values denote task accuracy or success rate (\%).
}
\label{tab:main-results}

\begin{tabular}{lcccc}
\toprule
Method &
SearchQA $\uparrow$ &
ALFWorld $\uparrow$ &
WebShop $\uparrow$ &
Avg. $\uparrow$ \\
\midrule

\multicolumn{5}{l}{\textit{Base Model}} \\
Qwen3.5-4B
& 58.1 & 32.1 & 10.8 & 33.7 \\

\midrule
\multicolumn{5}{l}{\textit{Explicit-State Evolution}} \\
SkillOpt-Lite
& 59.7 & 58.6 & 30.4 & 49.6 \\
SkillOpt
& 61.1 & 64.8 & 35.9 & 53.9 \\

\midrule
\multicolumn{5}{l}{\textit{Policy Internalization}} \\
SKILL0
& 60.6 & 61.5 & 31.8 & 51.3 \\
SkillC
& 61.4 & 64.2 & 36.8 & 54.1 \\
OPID
& 61.8 & 65.3 & 39.1 & 55.4 \\

\midrule
\multicolumn{5}{l}{\textit{State--Policy Co-Evolution}} \\
SkillRL
& 61.9 & 66.4 & 40.2 & 56.2 \\

\midrule
\multicolumn{5}{l}{\textit{Ours}} \\
\textbf{State--Policy Consolidation Loop}
& \textbf{62.4}
& \textbf{67.9}
& \textbf{42.4}
& \textbf{57.6} \\

\bottomrule
\end{tabular}
\end{table*}

The improvement is consistent across different adaptation paradigms.
Compared with the strongest explicit-state method, SkillOpt, our method improves
the average score from 53.9\% to 57.6\%. It also outperforms OPID, the strongest
policy-internalization baseline, by 2.2 points on average. More importantly,
our method surpasses SkillRL, the closest state--policy co-evolution baseline,
by 1.4 points on average, with gains on all three environments. These results
show that the advantage of our framework extends beyond either accumulating
explicit experience or internalizing it into the policy alone. Explicitly
coordinating state evolution, policy consolidation, and subsequent state
revision provides a stronger mechanism for continued agent improvement.

\begin{figure}[htbp]
    \centering
    \includegraphics[width=0.98\linewidth]
    {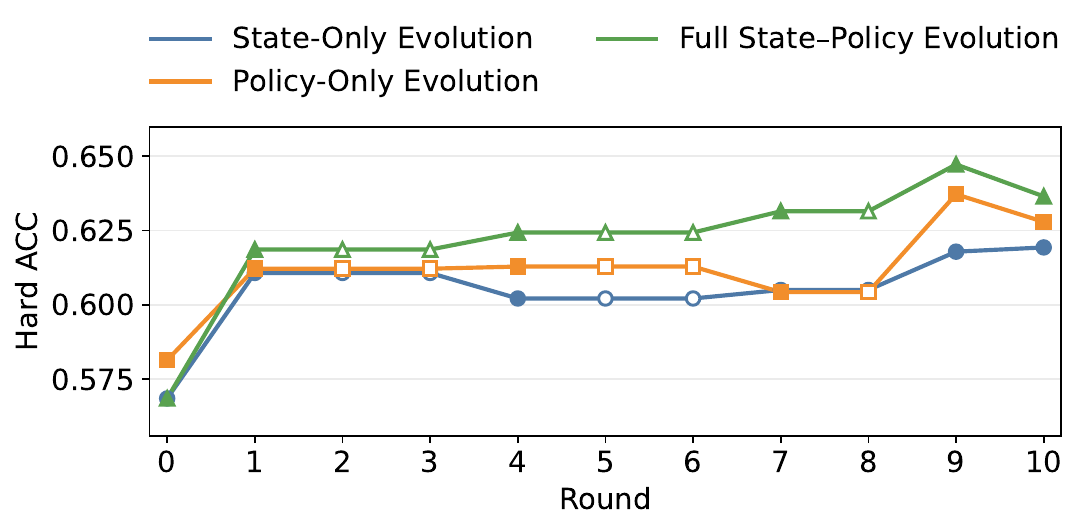}
    \caption{
    State--policy evolution on SearchQA.
    \textbf{Consolidated Policy (No State)} evaluates autonomous policy
    performance after consolidation.
    \textbf{Frozen Policy + Evolved State} evaluates each evolved state with
    the initial policy.
    \textbf{Matched State--Policy Pair} evaluates each evolved state with its
    contemporaneous policy.
    Filled markers denote accepted evolution rounds, while rejected rounds
    retain the preceding accepted checkpoint.
    }
    \label{fig:searchqa-state-policy-evolution}
\end{figure}

To further examine how this improvement emerges over successive evolution
rounds, Figure~\ref{fig:searchqa-state-policy-evolution} tracks policy and
state--policy performance on SearchQA. The state-free trajectory measures how
much experience has been absorbed into the policy, while the matched
state--policy trajectory captures the additional benefit of combining the
updated policy with its evolved explicit state.

Across accepted consolidation rounds, state-free policy accuracy increases
from 58.1\% to 61.3\%, demonstrating that interaction experience is
progressively converted into autonomous policy competence. Combining the final
consolidated policy with its evolved state further raises performance to
62.4\%. In contrast, pairing the same evolved state with the initial policy
reaches only 60.2\%, showing that the value of explicit experience increasingly
depends on the policy with which it evolves.

Taken together, the main results and evolution trajectories support the central
design of our framework. Policy consolidation continuously absorbs reusable
experience into model parameters, while state evolution preserves complementary
guidance that remains useful to the current policy. Their coordinated evolution
therefore yields stronger performance than either state-centric or
policy-centric adaptation alone, while also outperforming existing joint
evolution approaches.

\subsection{Ablation Study}
\label{sec:ablation}

We conduct component ablations on SearchQA to examine the individual
and combined contributions of state evolution and policy consolidation.

\begin{table}[t]
\centering
\scriptsize
\setlength{\tabcolsep}{3.5pt}
\caption{Component ablation on SearchQA.}
\label{tab:ablation}
\begin{tabular}{lccc}
\toprule
Method &
State Evol. &
Policy Consol. &
ACC $\uparrow$ \\
\midrule
State-Only
& \checkmark & -- & 61.1 \\
Policy-Only
& -- & \checkmark & 62.8 \\
Full State--Policy Loop
& \checkmark & \checkmark & \textbf{63.6} \\
\bottomrule
\end{tabular}
\end{table}

Both adaptation mechanisms improve agent performance independently.
State-Only Evolution reaches 61.1\% ACC, whereas Policy-Only
Consolidation achieves 62.8\%, suggesting that parametric
internalization provides a stronger standalone adaptation mechanism on
SearchQA. Combining both yields the best performance of 63.6\%,
outperforming State-Only and Policy-Only by 2.5 and 0.8 points,
respectively. These results indicate that explicit state evolution and
policy consolidation provide complementary benefits: consolidation
absorbs reusable experience into the policy, while the evolving state
retains additional guidance that remains useful beyond parametric
adaptation alone.

\subsection{Analysis of State--Policy Consolidation}

\subsubsection{Dynamic State--Policy Coupling}

We first examine how the compatibility between explicit state and policy
changes throughout evolution. For each stage, we compare the co-evolved
state--policy pair with a control that applies the same evolved state to
the initial policy. This comparison directly measures whether state
evolution becomes increasingly aligned with the policy being optimized.

\begin{table}[t]
\centering
\scriptsize
\setlength{\tabcolsep}{4pt}
\renewcommand{\arraystretch}{1.05}
\caption{
State--policy compatibility across evolution stages on SearchQA.
Each evolved state is evaluated with either its co-evolved policy or the
initial policy.
}
\label{tab:searchqa-coupling}

\begin{tabular}{lcc}
\toprule
Stage &
Co-Evolved Pair &
Initial-Policy Pair \\
&
ACC (\%) $\uparrow$ &
ACC (\%) $\uparrow$ \\
\midrule

Initial
& 56.9
& 56.9 \\

Intermediate
& 61.9
& 61.1 \\

Final
& \textbf{62.4}
& 60.2 \\

\bottomrule
\end{tabular}
\end{table}

The advantage of the co-evolved state--policy pair becomes more
pronounced as evolution proceeds. At the intermediate stage, pairing the
evolved state with its corresponding policy improves ACC from 61.1\% to
61.9\%. At the final stage, this gap expands to 2.2 points, with the
co-evolved pair reaching 62.4\% compared with 60.2\% when the same state
is paired with the initial policy. These results show that the benefit
of explicit experience is increasingly tied to the policy with which it
evolves. The growing compatibility gap provides direct empirical support
for our central design choice of jointly evolving state and policy rather
than treating explicit experience as policy-independent guidance.

\subsubsection{Cross-Version Experience Selection}

We next investigate which evolving experiences are most useful for policy
consolidation. Our cross-version analysis distinguishes \emph{newly
useful} ($01$) experiences, which become beneficial only after the state
update, from \emph{persistently useful} ($11$) experiences, which remain
beneficial across successive state versions. We compare these
transition-specific subsets with unfiltered training and their combined
selection ($01+11$).

\begin{figure}[htbp]
    \centering
    \includegraphics[width=0.98\linewidth]
    {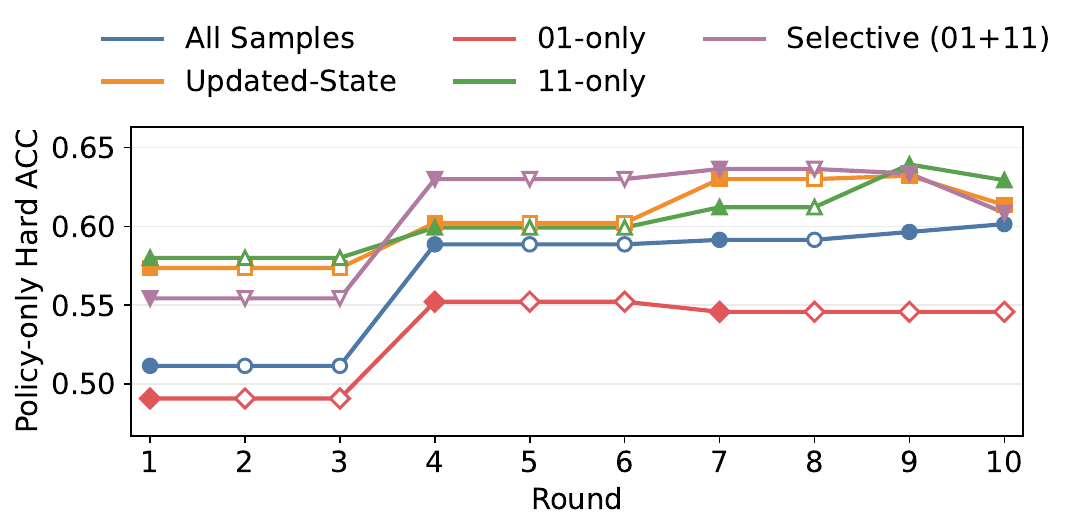}
    \caption{
    State-free policy accuracy under different experience-selection
    strategies across consolidation rounds.
    }
    \label{fig:searchqa-selection-roundwise}
\end{figure}

After the second consolidation round, combining newly useful and
persistently useful experiences achieves the best performance, reaching
63.0\% ACC. In comparison, unfiltered training reaches only 58.9\%,
while using $01$ or $11$ alone obtains 55.2\% and 59.9\%, respectively.
The $01+11$ strategy therefore improves over unfiltered consolidation by
4.1 points and over the stronger single-transition strategy by 3.1
points.

These results highlight the benefit of selectively consolidating
experience according to how its utility evolves across state versions.
Neither newly acquired nor persistent experience alone captures all of
the useful supervision: their combination provides a substantially
stronger learning signal for the state-free policy. This supports the
core motivation of our cross-version formulation---policy consolidation
should focus on experience that remains behaviorally useful during state
evolution rather than indiscriminately distilling all available
trajectories.

Under the current binary transition definition, the $01+11$ subset is
equivalent to the set of examples that are successful under the updated
state; we therefore use the comparison here to establish the benefit of
selective consolidation, while analyzing transition-specific credit
separately.

\subsubsection{Teacher Capacity and State-Conditioned Transfer}

The policy improvement could in principle arise from generic knowledge transferred by the larger teacher rather than from the explicit state.
To separate these effects, we vary both teacher scale and state conditioning while keeping the student initialization, selected training examples, OPD schedule, and evaluation split fixed.

\begin{table}[htbp]
    \centering
    \small
    \caption{
    Effect of teacher capacity and state conditioning on SearchQA.
    ACC measures the resulting state-free student policy.
    }
    \label{tab:searchqa-teacher}
    \begin{tabular}{llc}
        \toprule
        Teacher & Teacher Context & ACC $\uparrow$ \\
        \midrule
        No Teacher & No State & 58.14 \\
        4B Teacher & No State & 59.79 \\
        4B Teacher & Updated State & 57.57 \\
        27B Teacher & No State & 56.29 \\
        27B Teacher & Updated State & \textbf{61.21} \\
        \bottomrule
    \end{tabular}
\end{table}

The 27B teacher improves the student only when conditioned on the updated state.
Without state conditioning, the same teacher produces 0.5629 ACC, below the 0.5814 base policy, whereas conditioning it on the evolved state yields the strongest result of 0.6121.
The gain therefore cannot be explained by teacher scale alone.
Instead, the result suggests an interaction between teacher capacity and explicit experience: the larger teacher is effective when it serves as an executor of the evolved state rather than as a generic distillation source.

The 4B teacher does not show the same benefit from state conditioning.
This observation is consistent with our use of an asymmetric teacher--student configuration, although repeated runs are required to determine whether the interaction is statistically reliable.

\subsubsection{Residual State after Policy Consolidation}

We further examine how policy consolidation changes the role of explicit
state at deployment. A desirable state--policy pair should internalize
reusable experience into the policy while retaining explicit guidance
only when it remains complementary.

\begin{table}[htbp]
    \centering
    \small
    \caption{
    SearchQA deployment performance before and after policy consolidation.
    }
    \label{tab:searchqa-residual}
    \begin{tabular}{lc}
        \toprule
        Deployment Configuration &
        ACC (\%) $\uparrow$ \\
        \midrule

        Initial Policy \textit{w/o} State
        & 58.1 \\

        Consolidated Policy \textit{w/o} State
        & 61.3 \\

        Consolidated Policy \textit{w/} Full State
        & \textbf{62.4} \\

        Consolidated Policy \textit{w/} Residual State
        & 61.3 \\

        \bottomrule
    \end{tabular}
\end{table}

As shown in Table~\ref{tab:searchqa-residual}, policy consolidation
substantially improves state-free performance from 58.1\% to 61.3\%,
demonstrating that the evolving policy successfully absorbs a large
portion of the useful experience originally carried by the explicit
state. Reintroducing the full evolved state further increases accuracy
to 62.4\%, indicating that a small amount of complementary information
remains external to the policy after consolidation.

Importantly, the residualized configuration maintains the 61.3\%
performance of the consolidated policy without requiring the complete
evolved state. Together, these results illustrate the intended behavior
of our State--Policy Consolidation Loop: reusable experience is
progressively transferred into parametric competence, while explicit
state is reserved only for information that remains useful beyond what
the policy has already internalized.

\section{Related Work}

\subsection{Self-Evolving Agents}

A growing body of work enables language agents to improve from their own
interaction experience by maintaining and revising explicit textual knowledge.
Early approaches externalize feedback and successful behaviors as reflections,
episodic experience, or reusable skills that can guide subsequent interactions
without modifying the underlying model parameters
\cite{shinn2023reflexion,zhao2024expel,wang2024voyager}.
More recent work has broadened this paradigm toward persistent agent evolution.
ELL formalizes experience-driven lifelong learning around experience exploration,
long-term memory, skill learning, and knowledge internalization
\cite{cai2025ell}, while ACE treats agent context as an evolving playbook that is
incrementally generated, reflected upon, and curated from execution feedback
\cite{zhang2025ace}. SkillOpt further formulates a textual skill as the external
state of a frozen agent and optimizes it through bounded edits and held-out
validation, providing a controlled mechanism for text-space skill evolution
\cite{yang2026skillopt}. These approaches demonstrate that editable textual
state can serve as an effective substrate for rapid post-deployment adaptation.

Recent methods increasingly recognize that external state should evolve together
with the policy that executes it. SkillRL recursively updates a hierarchical
skill library alongside reinforcement learning, allowing reusable strategies to
co-evolve with policy behavior \cite{xia2026skillrl}. ReSkill makes this coupling
more explicit by evaluating competing skill revisions under the evolving policy
and continuously creating, testing, refining, and pruning skills according to
their utility \cite{he2026reskill}. At the broader agent-system level,
HarnessForge represents an agent as a harness--policy pair and jointly adapts
execution structure and policy behavior through harness tailoring and
harness-conditioned policy alignment \cite{chen2026harnessforge}. Recursive
Harness Self-Improvement instead focuses on prompt-level harness evolution,
iteratively refining the harness using pairwise feedback over its own revision
history \cite{lee2026rhi}. Together, these works establish that textual guidance
and parametric policy behavior are tightly coupled rather than independently
optimizable components.

Our work builds on this perspective but focuses on a different aspect of
state--policy evolution: how accumulated experience moves between the two
representations across successive deployment rounds. Rather than optimizing only
the current textual scaffold or its compatibility with the current policy, we
maintain a shared textual state synthesized from recurring evidence across
multiple trajectories, selectively consolidate state-enabled behavior into the
policy, and use subsequent interactions with the updated policy to determine
which accumulated guidance should be retained, revised, or retired.

\subsection{Skill Distillation and Experience Internalization}

A complementary line of work seeks to convert externally represented experience
into autonomous parametric behavior. Self-Consolidation summarizes reusable
patterns from historical successes and failures and distills the resulting
non-parametric experience into learnable model parameters
\cite{yu2026selfconsolidation}. More generally, On-Policy Context Distillation
(OPCD) trains a policy on its own trajectories while matching a
context-conditioned teacher through reverse-KL distillation, demonstrating that
experiential knowledge and optimized prompts can be internalized into a
state-free model \cite{ye2026opcd}. SKILL0 addresses the same objective through a
curriculum that progressively withdraws external skills according to their
on-policy helpfulness, encouraging the policy to transition from
skill-conditioned execution toward autonomous behavior
\cite{lu2026skill0}. These methods establish experience internalization as an
alternative to indefinitely retaining external guidance at inference time.

More recent approaches develop finer-grained mechanisms for deciding how skill
supervision should affect policy learning. Skill-SD summarizes completed
trajectories into natural-language skills that are provided only to a
privileged teacher, while the plain-prompt student learns to reproduce the
skill-conditioned behavior through self-distillation
\cite{wang2026skillsd}. SkillC instead samples paired skill-conditioned and
skill-free rollouts and converts their task-level performance contrast into a
direct credit signal, distinguishing skill-dependent success from autonomous
success during policy optimization \cite{lin2026skillc}. OPID constructs
hierarchical hindsight skills from completed on-policy trajectories and
re-scores sampled actions under ordinary and skill-augmented contexts; the
resulting token-level probability shifts provide dense skill-attributed
supervision in addition to trajectory-level rewards \cite{yang2026opid}. SEED
further makes this process self-evolving by using the current policy both to
collect interaction trajectories and to analyze them into hindsight skills,
allowing skill supervision to evolve together with the policy
\cite{wu2026seed}. UCOB extends skill-conditioned self-distillation
bidirectionally, treating skill-conditioned and skill-free prompts as two
on-policy views and using their relative return to determine local supervision
as well as utility-aware skill-memory updates \cite{tu2026ucob}.

These methods progressively improve how the behavioral contribution of external
skills is attributed and internalized. However, their internalization signals
primarily characterize the utility or behavioral effect of a current skill
relative to skill-free execution. Our method additionally tracks how this utility
changes across successive versions of a persistent textual state. Specifically,
we contrast state-free, previous-state, and updated-state behavior to distinguish
newly useful, persistently useful, regressive, and inactive state effects.
These cross-version transitions determine which trajectories contribute to
policy consolidation, while token-level state-conditioned contrasts localize
the decisions to be distilled. The resulting policy then generates new
interaction evidence for the next round of state evolution, closing an iterative
state--policy consolidation loop.

\section{Conclusion}

We introduced the \emph{State--Policy Consolidation Loop}, a framework for
coordinating explicit experience evolution and parametric policy improvement in
self-evolving agents. Rather than treating textual skills as either persistent
inference-time guidance or one-shot distillation targets, our framework maintains
an evolving textual state that summarizes recurring interaction experience and
selectively consolidates useful state-enabled behavior into the policy.
\emph{Transition-Aware Skill Distillation} further identifies consolidation
targets at both the trajectory and token levels by tracking behavioral utility
across state-free, previous-state, and updated-state execution.

Across embodied interaction, web navigation, and execution-verified coding, our
experiments show that alternating state evolution and policy consolidation
improves autonomous policy competence over state-only and internalization-only
alternatives, while reducing persistent dependence on accumulated textual
guidance. As the policy evolves, subsequent interaction experience also provides
feedback for retaining, revising, or retiring explicit state, enabling experience
to remain aligned with the agent's changing capabilities. These results suggest
that effective self-evolution requires not only acquiring more experience, but
continually determining what should remain explicit and what should become
parametric competence.

\section*{Limitations}

Our framework is primarily designed for post-deployment settings in which an
agent repeatedly encounters related tasks and receives sufficiently informative
environment-level feedback. This assumption makes the framework particularly
well suited to interactive and verifiable domains, such as web navigation,
tool use, and other recurring workflows, but may limit its direct applicability
to one-shot tasks or domains with highly subjective and sparse feedback.
Extending state--policy consolidation to such settings may require alternative
utility signals, such as learned verifiers or preference-based feedback.

The iterative evolution process also incurs additional computation for
multi-condition rollout evaluation, state revision, and policy consolidation.
This represents a compute--adaptation trade-off: additional evolution-time
computation is used to obtain a stronger state-free policy and reduce repeated
dependence on accumulated textual guidance. Importantly, these updates need not
be performed on the latency-critical serving path. Interaction experience can
be accumulated online and consolidated periodically or asynchronously during
lower-utilization periods, consistent with recent online--offline consolidation
paradigms \cite{lin2025sleep,ye2026autodreamer,zhang2026lightmem}. Such
scheduling reduces the impact on serving latency, although it introduces a
delay between experience acquisition and model updates. Improving update
scheduling, rollout reuse, and consolidation efficiency therefore remains an
important direction for long-term deployment.

\bibliography{custom}

@misc{shinn2023reflexion,
  title={Reflexion: Language Agents with Verbal Reinforcement Learning},
  author={Shinn, Noah and Cassano, Federico and Berman, Edward and Gopinath, Ashwin and Narasimhan, Karthik and Yao, Shunyu},
  year={2023},
  eprint={2303.11366},
  archivePrefix={arXiv},
  primaryClass={cs.AI},
  url={https://arxiv.org/abs/2303.11366}
}

@inproceedings{zhao2024expel,
  title={ExpeL: LLM Agents Are Experiential Learners},
  author={Zhao, Andrew and Huang, Daniel and Xu, Quentin and Lin, Matthieu and Liu, Yong-Jin and Huang, Gao},
  booktitle={Proceedings of the AAAI Conference on Artificial Intelligence},
  year={2024},
  volume={38},
  number={17},
  pages={19632--19642},
  doi={10.1609/AAAI.V38I17.29936}
}

@article{wang2024voyager,
  title={Voyager: An Open-Ended Embodied Agent with Large Language Models},
  author={Wang, Guanzhi and Xie, Yuqi and Jiang, Yunfan and Mandlekar, Ajay and Xiao, Chaowei and Zhu, Yuke and Fan, Linxi and Anandkumar, Anima},
  journal={arXiv preprint arXiv:2305.16291},
  year={2023},
  eprint={2305.16291},
  archivePrefix={arXiv},
  primaryClass={cs.AI},
  url={https://arxiv.org/abs/2305.16291}
}

@article{cai2025ell,
  title={Building Self-Evolving Agents via Experience-Driven Lifelong Learning: A Framework and Benchmark},
  author={Cai, Yuxuan and Hao, Yipeng and Zhou, Jie and Yan, Hang and Lei, Zhikai and Zhen, Rui and Han, Zhenhua and Yang, Yutao and Li, Junsong and Pan, Qianjun and Huai, Tianyu and Chen, Qin and Li, Xin and Chen, Kai and Zhang, Bo and Qiu, Xipeng and He, Liang},
  journal={arXiv preprint arXiv:2508.19005},
  year={2025},
  eprint={2508.19005},
  archivePrefix={arXiv},
  primaryClass={cs.AI},
  url={https://arxiv.org/abs/2508.19005}
}

@article{zhang2025ace,
  title={Agentic Context Engineering: Evolving Contexts for Self-Improving Language Models},
  author={Zhang, Qizheng and Hu, Changran and others},
  journal={arXiv preprint arXiv:2510.04618},
  year={2025},
  eprint={2510.04618},
  archivePrefix={arXiv},
  primaryClass={cs.AI},
  url={https://arxiv.org/abs/2510.04618}
}

@article{yang2026skillopt,
  title={SkillOpt: Executive Strategy for Self-Evolving Agent Skills},
  author={Yang, Yifan and Gong, Ziyang and Huang, Weiquan and Yang, Qihao and Zhou, Ziwei and Huang, Zisu and Li, Yan and Gao, Xuemei and Dai, Qi and Liu, Bei and Qiu, Kai and Yang, Yuqing and Chen, Dongdong and Yang, Xue and Luo, Chong},
  journal={arXiv preprint arXiv:2605.23904},
  year={2026},
  eprint={2605.23904},
  archivePrefix={arXiv},
  primaryClass={cs.AI},
  url={https://arxiv.org/abs/2605.23904}
}

@article{shen2026skilloptlite,
  title={SkillOpt-Lite: Better and Faster Agent Self-evolution via One Line of Vibe},
  author={Shen, Yifei and Li, Bo and Zhang, Xinjie},
  journal={arXiv preprint arXiv:2607.03451},
  year={2026},
  eprint={2607.03451},
  archivePrefix={arXiv},
  primaryClass={cs.SE},
  url={https://arxiv.org/abs/2607.03451}
}

@article{zhang2026memskill,
  title={MemSkill: Learning and Evolving Memory Skills for Self-Evolving Agents},
  author={Zhang, Haozhen and Long, Quanyu and Bao, Jianzhu and Feng, Tao and Zhang, Weizhi and Yue, Haodong and Wang, Wenya},
  journal={arXiv preprint arXiv:2602.02474},
  year={2026},
  eprint={2602.02474},
  archivePrefix={arXiv},
  primaryClass={cs.AI},
  url={https://arxiv.org/abs/2602.02474}
}

@article{yu2026selfconsolidation,
  title={Self-Consolidation for Self-Evolving Agents},
  author={Yu, Hongzhuo and Zhu, Fei and Xie, Guo-Sen and Shao, Ling},
  journal={arXiv preprint arXiv:2602.01966},
  year={2026},
  eprint={2602.01966},
  archivePrefix={arXiv},
  primaryClass={cs.LG},
  url={https://arxiv.org/abs/2602.01966}
}

@article{lu2026skill0,
  title={SKILL0: In-Context Agentic Reinforcement Learning for Skill Internalization},
  author={Lu, Zhengxi and Yao, Zhiyuan and Wu, Jinyang and Han, Chengcheng and Gu, Qi and Cai, Xunliang and Lu, Weiming and Xiao, Jun and Zhuang, Yueting and Shen, Yongliang},
  journal={arXiv preprint arXiv:2604.02268},
  year={2026},
  eprint={2604.02268},
  archivePrefix={arXiv},
  primaryClass={cs.LG},
  url={https://arxiv.org/abs/2604.02268}
}

@article{liu2024lostmiddle,
  title={Lost in the Middle: How Language Models Use Long Contexts},
  author={Liu, Nelson F. and Lin, Kevin and Hewitt, John and Paranjape, Ashwin and Bevilacqua, Michele and Petroni, Fabio and Liang, Percy},
  journal={Transactions of the Association for Computational Linguistics},
  volume={12},
  pages={157--173},
  year={2024},
  doi={10.1162/tacl_a_00638}
}

@article{lin2026museautoskill,
  title={MUSE-Autoskill: Self-Evolving Agents via Skill Creation, Memory, Management, and Evaluation},
  author={Lin, Huawei and Li, Peng and Song, Jie and Jiang, Fuxin and Zhang, Tieying},
  journal={arXiv preprint arXiv:2605.27366},
  year={2026},
  eprint={2605.27366},
  archivePrefix={arXiv},
  primaryClass={cs.AI},
  url={https://arxiv.org/abs/2605.27366}
}

@article{xu2026hyperskill,
  title={HyperSkill: Self-Evolving LLM Agents via Hypergraph-Structured Skill Memory},
  author={Xu, Ruiyao and Yang, Tiankai and Huang, Wei-Chieh},
  journal={arXiv preprint arXiv:2608.16114},
  year={2026},
  eprint={2608.16114},
  archivePrefix={arXiv},
  primaryClass={cs.CL},
  url={https://arxiv.org/abs/2608.16114}
}

@article{xia2026skillrl,
  title={SkillRL: Evolving Agents via Recursive Skill-Augmented Reinforcement Learning},
  author={Xia, Peng and Chen, Jianwen and Wang, Hanyang and Liu, Jiaqi and Zeng, Kaide and Wang, Yu and Han, Siwei and Zhou, Yiyang and Zhao, Xujiang and Chen, Haifeng and others},
  journal={arXiv preprint arXiv:2602.08234},
  year={2026},
  eprint={2602.08234},
  archivePrefix={arXiv},
  primaryClass={cs.AI},
  url={https://arxiv.org/abs/2602.08234}
}

@article{he2026reskill,
  title={ReSkill: Reconciling Skill Creation with Policy Optimization in Agentic RL},
  author={He, Zelin and Lin, Haotian and Han, Boran and Zhu, Wei and Fang, Haoyang and Wang, Bernie and Zhu, Xuan and Li, Runze and Reimherr, Matthew},
  journal={arXiv preprint arXiv:2606.01619},
  year={2026},
  eprint={2606.01619},
  archivePrefix={arXiv},
  primaryClass={cs.AI},
  url={https://arxiv.org/abs/2606.01619}
}

@article{chen2026harnessforge,
  title={HarnessForge: Joint Harness and Policy Evolution for Adaptive Agent Systems},
  author={Chen, Mingju and Lv, Can and Zhang, Guibin and Chang, Heng and Zhou, Shiji},
  journal={arXiv preprint arXiv:2606.01779},
  year={2026},
  eprint={2606.01779},
  archivePrefix={arXiv},
  primaryClass={cs.CL},
  url={https://arxiv.org/abs/2606.01779}
}

@article{lee2026rhi,
  title={Recursive Harness Self-Improvement},
  author={Lee, Hyunin and Xu, Jinglue and Seely, Jeffrey and Lee, Donghyun and Zaharia, Matei and Tang, Yujin},
  journal={arXiv preprint arXiv:2607.15524},
  year={2026},
  eprint={2607.15524},
  archivePrefix={arXiv},
  primaryClass={cs.LG},
  url={https://arxiv.org/abs/2607.15524}
}

@article{wang2026skillsd,
  title={Skill-SD: Skill-Conditioned Self-Distillation for Multi-turn LLM Agents},
  author={Wang, Hao and Wang, Guozhi and Xiao, Han and Zhou, Yufeng and Pan, Yue and Wang, Jichao and Xu, Ke and Wen, Yafei and Ruan, Xiaohu and Chen, Xiaoxin and Qi, Honggang},
  journal={arXiv preprint arXiv:2604.10674},
  year={2026},
  eprint={2604.10674},
  archivePrefix={arXiv},
  primaryClass={cs.LG},
  url={https://arxiv.org/abs/2604.10674}
}

@article{lin2026skillc,
  title={SKILLC: Learning Autonomous Skill Internalization in LLM Agents via Contrastive Credit Assignment},
  author={Lin, Hongxiang and Kuai, Zhirui and Xue, Erpeng and Wang, Lei},
  journal={arXiv preprint arXiv:2605.27899},
  year={2026},
  eprint={2605.27899},
  archivePrefix={arXiv},
  primaryClass={cs.LG},
  url={https://arxiv.org/abs/2605.27899}
}

@article{yang2026opid,
  title={OPID: On-Policy Skill Distillation for Agentic Reinforcement Learning},
  author={Yang, Shuo and Wu, Jinyang and Lu, Zhengxi and Shen, Yuhao and Zhang, Fan and Feng, Lang and Zhang, Shuai and Luo, Haoran and Lian, Zheng and Wen, Zhengqi and Tao, Jianhua},
  journal={arXiv preprint arXiv:2606.26790},
  year={2026},
  eprint={2606.26790},
  archivePrefix={arXiv},
  primaryClass={cs.CL},
  url={https://arxiv.org/abs/2606.26790}
}

@article{wu2026seed,
  title={SEED: Self-Evolving On-Policy Distillation for Agentic Reinforcement Learning},
  author={Wu, Jinyang and Yang, Shuo and Lu, Zhengxi and Zhang, Fan and Shen, Yuhao and Feng, Lang and Luo, Haoran and Lian, Zheng and Zhang, Shuai and Wen, Zhengqi and Tao, Jianhua},
  journal={arXiv preprint arXiv:2607.14777},
  year={2026},
  eprint={2607.14777},
  archivePrefix={arXiv},
  primaryClass={cs.LG},
  url={https://arxiv.org/abs/2607.14777}
}

@article{ye2026opcd,
  title={On-Policy Context Distillation for Language Models},
  author={Ye, Tianzhu and Dong, Li and Wu, Xun and Huang, Shaohan and Wei, Furu},
  journal={arXiv preprint arXiv:2602.12275},
  year={2026},
  eprint={2602.12275},
  archivePrefix={arXiv},
  primaryClass={cs.CL},
  url={https://arxiv.org/abs/2602.12275}
}

@article{tu2026ucob,
  title={UCOB: Learning to Utilize and Evolve Agentic Skills via Credit-Aware On-Policy Bidirectional Self-Distillation},
  author={Tu, Songjun and Xu, Chengdong and Zhang, Qichao and Ma, Yiwen and Zhang, Yaocheng and Li, Linjing and Li, Dong and Lan, Xiangyuan and Zhao, Dongbin},
  journal={arXiv preprint arXiv:2606.29502},
  year={2026},
  eprint={2606.29502},
  archivePrefix={arXiv},
  primaryClass={cs.AI},
  url={https://arxiv.org/abs/2606.29502}
}

@article{lu2026sdar,
  title={Self-Distilled Agentic Reinforcement Learning},
  author={Lu, Zhengxi and Yao, Zhiyuan and Han, Zhuowen and Wang, Zi-Han and Wu, Jinyang and Gu, Qi and Cai, Xunliang and Lu, Weiming and Xiao, Jun and Zhuang, Yueting and Shen, Yongliang},
  journal={arXiv preprint arXiv:2605.15155},
  year={2026},
  eprint={2605.15155},
  archivePrefix={arXiv},
  primaryClass={cs.LG},
  url={https://arxiv.org/abs/2605.15155}
}

@article{lin2025sleep,
  title={Sleep-time Compute: Beyond Inference Scaling at Test-time},
  author={Lin, Kevin and Snell, Charlie and Wang, Yu and Packer, Charles and Wooders, Sarah and Stoica, Ion and Gonzalez, Joseph E.},
  journal={arXiv preprint arXiv:2504.13171},
  year={2025},
  eprint={2504.13171},
  archivePrefix={arXiv},
  primaryClass={cs.AI},
  url={https://arxiv.org/abs/2504.13171}
}

@article{ye2026autodreamer,
  title={Auto-Dreamer: Learning Offline Memory Consolidation for Language Agents},
  author={Ye, Chongrui and Liu, Yuxiang and Wang, Yu and Yu, Haofei and Zhao, Yining and Liu, Ge and McAuley, Julian and You, Jiaxuan},
  journal={arXiv preprint arXiv:2605.20616},
  year={2026},
  eprint={2605.20616},
  archivePrefix={arXiv},
  primaryClass={cs.CL},
  url={https://arxiv.org/abs/2605.20616}
}

@article{zhang2026lightmem,
  title={Lightweight LLM Agent Memory with Small Language Models},
  author={Zhang, Jiaquan and Zhang, Chaoning and Chen, Shuxu and Huang, Zhenzhen and Zheng, Pengcheng and Wang, Zhicheng and Guo, Ping and Mo, Fan and Bae, Sung-Ho and Zou, Jie and Wei, Jiwei and Yang, Yang},
  journal={Proceedings of the 64th Annual Meeting of the Association for Computational Linguistics},
  pages={12914--12929},
  year={2026},
  doi={10.18653/v1/2026.acl-long.588},
  url={https://aclanthology.org/2026.acl-long.588/}
}

\newpage
\appendix

\end{document}